 \documentclass[a4paper]{spie}  
\usepackage[]{graphicx}
\usepackage[hang]{subfigure}

\usepackage{fancyheadings}

\title{Comparing Multi-Target Trackers\\ on Different Force Unit Levels} 
\author{Hedvig Sidenbladh, Pontus Svenson and Johan Schubert
\skiplinehalf
Department of Data and Information Fusion\\ 
Division of Command and Control Systems\\
Swedish Defence Research Agency\\ SE--172 90 Stockholm, Sweden\\
{\tt hedvig,ponsve,schubert@foi.se}\\
{\tt http://www.foi.se/fusion/}
}

\begin{document} 

\thispagestyle{plain}

\maketitle

\begin{abstract}

Consider the problem of tracking a set of moving targets. Apart from the 
tracking result, it is often important to know where the tracking 
fails, either to steer sensors to that part of the state-space, or to inform a 
human operator about the status and quality of the obtained information.
An intuitive quality measure is the correlation 
between two tracking results based on uncorrelated observations. In the case 
of Bayesian trackers such a correlation measure could be the Kullback-Leibler 
difference.

We focus on a scenario with a large number of military units moving in 
some terrain. The units are observed by several types of sensors and 
"meta-sensors" with force aggregation capabilities. The sensors register 
units of different size. Two separate multi-target probability hypothesis density (PHD) particle filters are 
used to track some type of units (e.g., companies) and their sub-units 
(e.g., platoons), respectively, based on observations of 
units of those sizes. Each observation is used in one filter only.

Although the state-space may well be the same in both filters, the posterior 
PHD distributions are not directly comparable -- one unit might 
correspond to three or four spatially distributed sub-units. Therefore, we 
introduce a mapping function between distributions for different unit size, 
based on doctrine knowledge of unit configuration.

The mapped distributions can now be compared -- locally or globally -- using
some measure, which gives the correlation between two 
PHD distributions in a bounded volume of the state-space. To locate areas 
where the tracking fails, a discretized quality map of the 
state-space can be generated by applying the measure locally to different 
parts of the space.  
\end{abstract}

\section{Introduction}
\label{sec:intro}

Information fusion 
is the process of extracting meaningful
information from a large number of sources.
Such sources could be sensors of a wide variety of
types, but also includes pre-processed data on,
e.g., terrain, enemy doctrine and objectives.
Manual fusion has always been performed in
military staffs. However, with the increasing amount of data
provided by modern sensors, it is necessary
to automate the process as much as possible. 

Due to sensor noise and model deficiencies, automatic information
fusion methods always provide an approximative view of the situation. 
The degree of approximation indicates to what degree the fused 
information can be trusted. 
Therefore, to make an information fusion system useful in practice, it 
is very important to develop methods for measuring the degree of 
approximation or method failure, either to inform a human operator of the 
situation or to provide input to a sensor management module. 

Using several independent methods for solving the
same subproblem in the fusion system enhances the reliability of the system
in two aspects: firstly, if one method fails, the other may succeed; secondly,
the outputs of the different methods can be compared -- similarity of results
indicates that both methods are functioning. 

\begin{figure}[t]
\centerline{\includegraphics[width=16.75cm]{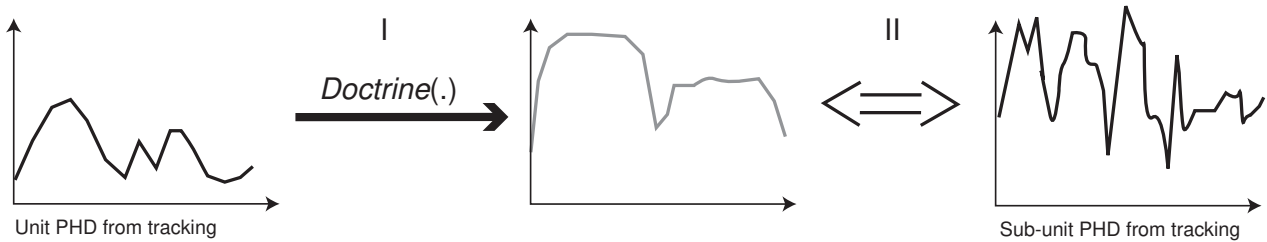}}
\caption{Overview of the method. 
I Estimating the sub-unit states from the unit states based on doctrine 
knowledge (Section \ref{sec:transform}). 
II Comparing PHD distributions (Section \ref{sec:compare}).}
\label{fig:overview}
\end{figure}

In this work, we will concentrate on the second point, and
provide an example of how
a signal or measure of quality could be used
to determine when two tracking methods
differ. In particular, we will assume the output of both tracking methods to 
be on the form of a probability hypothesis density (PHD) 
\cite{mahler-zajic01}. Our method can be schematically described as in 
Figure~\ref{fig:overview}. 
The input to the method are the two independently achieved PHD functions 
shown to the far left and right. 

We now assume that there exist some doctrine knowledge on how sub-units are 
spatially organized in units. This knowledge is implemented as an operator 
$\mathit{Doctrine}(\cdot)$ so that a ``synthesized'' sub-unit PHD $D^*_{SU}$ can 
be generated from the unit PHD $D_U$ as
\begin{equation}
 D^*_{SU} = \mathit{Doctrine}(D_U) ~.
\end{equation}
This is illustrated as step I in Figure~\ref{fig:overview}.
The ``synthesized'' PHD $D^*_{SU}$ can be directly compared with the 
actual sub-unit PHD $D_{SU}$ as shown in step II.

The output of the comparison method could be used
either to inform a human system operator, or to provide input to an
automatic sensor adaptation module.

In short, the contributions presented in this paper are
\begin{enumerate}
\item The idea of extracting a signal of quality online, based on 
      comparison of the output of two statistically independent fusion methods.
\item A method for estimating the PHD of the states of
      military sub-units from a unit PHD based on doctrine knowledge.
\end{enumerate}

The remainder of the paper is outlined in the following way. In Section 
\ref{sec:related} we discuss previous work on the subject. 
Section \ref{sec:problem} introduces the multi-target
tracking problem we have chosen to try our method on.
The tracking method is shortly introduced in Section \ref{sec:tracking}.
Transforming a unit PHD to a sub-unit PHD using doctrines is described in 
Section \ref{sec:transform}, while Section~\ref{sec:compare} describes the 
specific methods of comparing PHD functions that we have used in this paper. 
Results on a one-dimensional (1D) example scenario are presented in Section
\ref{sec:experiments}. The paper is concluded with a discussion and 
description of future work in Section \ref{sec:conclusions}.

\section{Related Work}
\label{sec:related}

In this paper, we describe measures of difference between PHD distributions.
If the output of our fusion methods would be on another form, other difference 
measures would be suitable. 

A straightforward measure of difference 
is the Dempster-Shafer~\cite{shaferbook} conflict
between two pieces of evidence. In case of large structural difference between 
two pieces of evidence, application of the standard rule of combination
will result in a large conflict. If the two pieces evidence are output of two
independent fusion modules solving the same problem,
then a large conflict means that at least one of them has been corrupted.

Mahler \cite{mahler98}, Zajic and Mahler \cite{zajicmahler99},and Hoffman et 
al.~\cite{hoffmanmahlerzajic01} have used generalized Kullback-Leibler 
difference (Section \ref{sec:compare}) and Csisz{\'a}r metrics to measure the 
efficiency and correctness of 
fusion methods in a number of papers. Although we have a similar goal, there 
are several differences between the approach taken in these papers and our 
work.

Firstly, the data compared in papers 
\cite{mahler98,zajicmahler99,hoffmanmahlerzajic01} is on the form of full 
random sets~\cite{goodman97}, while our data is on the form of PHD functions
(Section \ref{sec:tracking}). Obviously, the problem of defining a distance 
between two multi-target probability density functions over random sets is 
quite different from the problem of defining a distance metric between two 
PHD functions.

Secondly, our goal is to obtain an online quality measure based on the 
difference between the outputs of two fusion methods, while the goal of 
Mahler et al.~is to obtain a measure of the information content, i.e., an 
entropy measure, in the multi-target probability density function, or a 
measure of the difference between the obtained function and a known ground 
truth.

\section{The Problem}
\label{sec:problem}

The problem addressed in this paper is the following: A data fusion system
has two tracking methods at its disposal. One tracking method gives as output 
a PHD over the state of units (e.g., position, velocity, type, composition, 
goal), another gives a PHD over the state of sub-units, which are 
organized in units. The input to the two methods can be considered 
statistically independent given the model, since the observations originate 
from different types of sensors.

It should be noted that these sensors are to be regarded as ``meta-sensors'', 
in that they give observations of entire units. The 
observations could, e.g., be generated from soldiers or
civilians who are capable of recognizing units. They could
also be the output of an independent force aggregation
algorithm~\cite{schubert03}.

The consistency of the two methods is related 
to the quality of tracking -- a high correspondence between the trackers 
indicates good performance of both methods, while a low correspondence 
indicates that at least one of the trackers is failing given that the 
assumed doctrine (Section \ref{sec:transform}) is correct.
The goal of this work is thus to develop a method for comparing the two 
different PHD functions.

\section{PHD Particle Filtering}
\label{sec:tracking}

The tracking methods are instances of PHD particle filters 
\cite{sidenbladhFUSION03} which is a particle implementation of the PHD filter
\cite{mahler-zajic01}. 

A probability hypothesis density (PHD) is the first moment of the joint 
multi-target probability density over a random set of objects $\Gamma$ 
\cite{goodman97,mahler-zajic01}. 
It is defined over the state-space $\Theta$ of one object (here, one unit or 
sub-unit).
The integral of a PHD $D$ over any volume $S$ in the state-space is the 
expected number of objects in that volume,
\begin{equation}
E(|\Gamma \cap S|) = \int_S D(\mathbf{x})d\mathbf{x}
\end{equation}
where $|\cdot|$ denotes cardinality. Hence, the PHD can be intuitively regarded 
as an estimate of the
``object density'' in the state-space. It should be noted that the PHD is not 
a probability density, since the integral over $\Theta$ is not 1, but the 
estimated total number of objects.

A PHD filter propagates a PHD in time, incorporating new observations with the 
tracking history in a mathematically grounded manner
\cite{mahler-zajic01,sidenbladhFUSION03}.

\section{Transforming a unit PHD to a sub-unit PHD}
\label{sec:transform}

Military vehicles organized in units usually move in certain spatial 
configurations according to military doctrine. Here we make use of this 
knowledge to estimate a PHD over sub-unit states given a PHD over unit 
states.\footnote{It should be noted that this is a limitation of the method, 
as doctrine knowledge is of limited use in, e.g., OOTW applications.} 

For ease of visualization, we take a 1D example (where the state variable 
could be the position of units and their sub-units moving along a road). 
A doctrine could then state that a unit consists of three sub-units. The 
sub-units are evenly spread out, their center of gravity at the position of 
the unit. Furthermore, due to the human factor, the actual inter-unit 
distances deviate randomly from the distance decreed by the doctrine. 

\begin{figure}[t]
\centerline{\includegraphics[width=9cm]{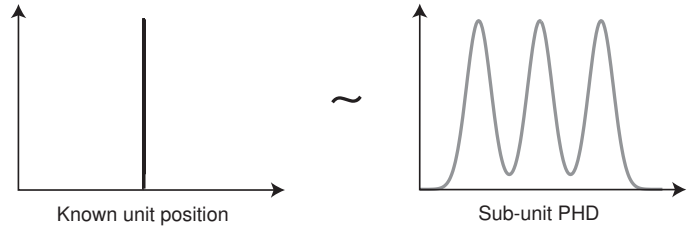}}
\caption{A doctrine expressed as a convolution mask -- exact knowledge of the 
unit state corresponds to a PHD over sub-unit state.}
\label{fig:platoon}
\end{figure}

This can be expressed as a PHD as shown in Figure \ref{fig:platoon}. Given 
that we know the unit position, the sub-unit positions can be modeled as a
PHD $D_{\mathit{Doctrine}}$
which is the sum of three normal distributions, centered in the three 
sub-unit positions according to doctrine, with a standard deviation 
corresponding to the expected random deviation from the doctrine.

Using this formulation of doctrine and random deviation from the doctrine, 
the operator $\mathit{Doctrine}(\cdot)$ discussed in the introduction can now 
be defined as
\begin{equation}
D^*_{SU} = \mathit{Doctrine}(D_U) \equiv D_{\mathit{Doctrine}} \otimes D_U
\label{eq:conv}
\end{equation}
where $\otimes$ denotes convolution and $D_{\mathit{Doctrine}}$ acts as a 
convolution mask.\footnote{The relevance of this can be 
tested by a simple experiment of thought. Compute 
$\mathit{Doctrine}(\delta)$ where $\delta$ is a Dirac 
pulse, corresponding to exact knowledge of the state of the unit. Since 
$D \otimes \delta = D$ for any density $D$, the result 
will be the PHD $D_{\mathit{Doctrine}}$, i.e., the sub-unit PHD that would 
result if the unit state is known.}
This procedure corresponds to step I in Figure~\ref{fig:overview}.

The method also generalizes to an $N$-dimensional state space; 
$D_{\mathit{Doctrine}}$ then becomes an $N$-dimensional PHD, which is 
convolved with $D_{U}$, also $N$-dimensional. 

In the case of several possible doctrines, two different approaches can be 
taken. The first alternative is to define $D_{\mathit{Doctrine}}$ as a
superposition of the individual doctrine PHD functions. The other alternative 
is to 
perform the whole comparison using each choice of doctrine, and then select 
the doctrine that displays the best match with the actual sub-unit PHD. The 
latter approach could also be used for doctrine recognition.

\section{Comparing PHD Distributions}
\label{sec:compare}

A common measure for comparison of two probability density functions $f$ and 
$g$ is the Kullback-Leibler divergence
\begin{equation}
K(f,g) = \int f(\mathbf{x}) \log{\frac{f(\mathbf{x})}{g(\mathbf{x})}} 
              d\mathbf{x} ~.
\label{eq:kl}
\end{equation}
However, this measure does not suit our purposes. There are two reasons for 
this. Firstly, the measure is only well defined when 
$\int f(\mathbf{x})d\mathbf{x} = \int g(\mathbf{x})d\mathbf{x}$, which is the 
case for probability density functions, but not for PHD functions. Secondly, 
it is not a 
proper metric, since $K(f,g) \neq K(g,f)$ in the general case. The divergence 
was defined to measure the difference between the ground truth $f$ and an 
approximation $g$ of the ground truth, or the entropy of a distribution $f$
compared to a non-informative, or prior, distribution $g$ - not the 
difference between two different approximations.

Instead, we use standard norms as a distance metric. The norm of the 
difference between two PHD functions gives a measure of the difference in the 
estimated number of objects, as well as the difference in estimated object 
state.

The $L_p$ norm of a function $f$ is defined as
\begin{equation}
\parallel f \parallel_p \, \equiv
(\int | f(\mathbf{x}) |^p d\mathbf{x} ) ^{1/p}
\label{eq:lnorm}
\end{equation}
where $|\cdot|$ denotes absolute value. For the special case of $p=\infty$,
we define
\begin{equation}
\parallel f \parallel_{\infty} \, \equiv \max {f} ~.
\end{equation}

Using this norm, a distance function between the two PHD functions $D_{SU}$ 
and $D^*_{SU}$ is defined as
\begin{equation}
d_p(D_{SU}, D^*_{SU}) = \, \parallel D_{SU} - D^*_{SU} \parallel_p ~.
\end{equation}
The distance measures $d_1(D_{SU}, D^*_{SU})$, $d_2(D_{SU}, D^*_{SU})$ and 
$d_\infty(D_{SU}, D^*_{SU})$ are computed
in Section~\ref{sec:experiments}.
The extraction of this distance corresponds to step II in 
Figure~\ref{fig:overview}.

\section{Implementation of the method}
\label{sec:experiments}

To illustrate the idea, the method of comparison is implemented with a 
simulated 1D scenario, outlined below.

\subsection{Scenario}
\label{sec:experiments_scenario}
In the scenario, a unit moves in a 1D state-space (e.g., along a road) with a 
velocity $v_t$. Here, $v_t$ is sampled in each time-step $t$ from a normal 
distribution $N(v_U, \sigma_U)$. Let $x_t$ be the position of the unit at 
time $t$.

The unit consists of three sub-units. Their positions $x^1_t$, $x^2_t$ and 
$x^3_t$ are generated in each time-step $t$ from the doctrine discussed in 
Section \ref{sec:transform}. Let $x_{\mathit{Doctrine}}$ be the 
inter-sub-unit distance, and $\sigma_{\mathit{Doctrine}}$ the expected 
standard deviation from the doctrine. This gives the sub-unit positions
\begin{eqnarray} 
&&x^1_t = x_t - x_{\mathit{Doctrine}} + n^1\\
&&x^2_t = x_t + n^2\\
&&x^3_t = x_t + x_{\mathit{Doctrine}} + n^3
\end{eqnarray} 
where $n^1$, $n^2$ and $n^3$ are 
sampled from $N(0, \sigma_{\mathit{Doctrine}})$.
 
The input to the two PHD particle filters \cite{sidenbladhFUSION03} are 
observations of the ground truth positions and velocities. Each unit is 
observed with probability $p_{FN}$. An observation of a state $[x_t, v_t]$
of a unit (or sub-unit) is defined as
\begin{equation}
\mathbf{y}_t = [x_t, v_t] + \mathbf{n}_O
\end{equation}
where $\mathbf{n}_O$ is a noise term 
sampled from $N(0, \mathbf{\sigma}_O)$. The output from the filters are in 
each time-step $t$ the two (discrete) PHD functions $D_{U,t}$ and $D_{SU,t}$.

A convolution mask $D_{\mathit{Doctrine}}$ (Section \ref{sec:transform}) is 
constructed from the doctrine as 
\begin{equation}
D_{\mathit{Doctrine}} = 
  N(-x_{\mathit{Doctrine}}, \sigma_{\mathit{Doctrine}}) + 
  N(0, \sigma_{\mathit{Doctrine}}) + 
  N(x_{\mathit{Doctrine}}, \sigma_{\mathit{Doctrine}}) ~.
\end{equation}
This mask is used in every time-step $t$ to obtain the ``synthesized'' sub-unit 
PHD $D^*_{SU,t}$ according to Equation (\ref{eq:conv}).

\begin{figure}[t]
\centerline{\subfigure[Very exact doctrine.]{\includegraphics[width=13.63cm]{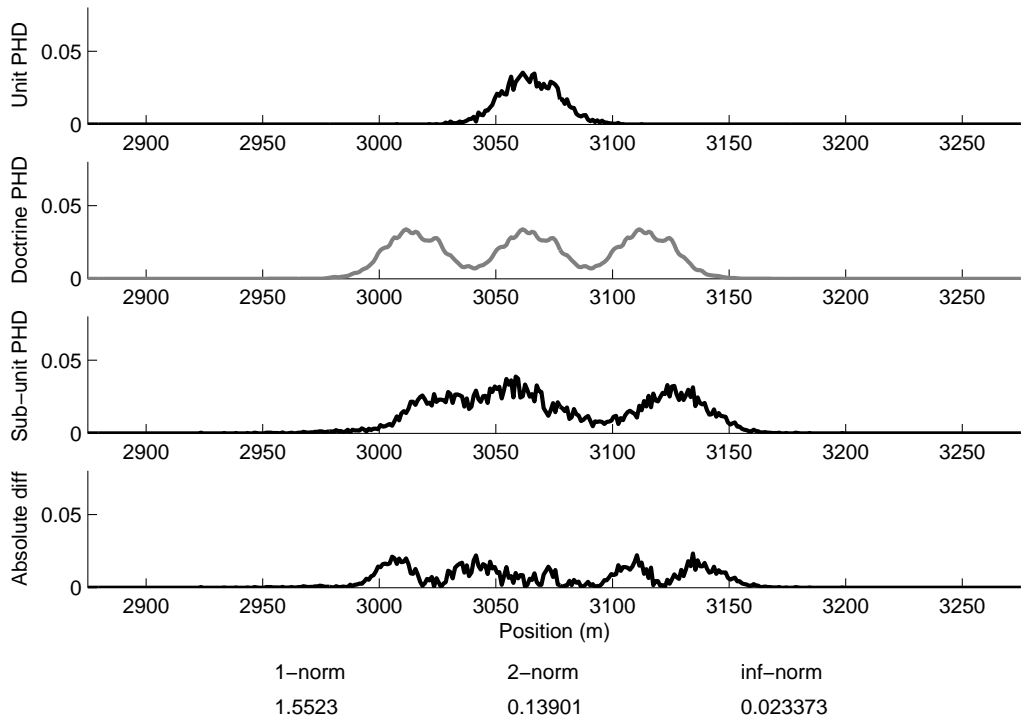}}}
\caption{Tracking in a 1D simulated scenario. (a) Very exact doctrine, 
$\sigma_{\mathit{Doctrine}} \rightarrow 0$. (b) Quite exact doctrine, 
$\sigma_{\mathit{Doctrine}} = \sigma^x_O$ {\em (On next page)}. (c) Very loose 
doctrine, $\sigma_{\mathit{Doctrine}} = x_{\mathit{Doctrine}}$ 
{\em (On next page)} .}
\label{fig:experiments}
\end{figure}

\begin{figure}[p]
\setcounter{subfigure}{1}
\centerline{\subfigure[Quite exact doctrine.]{\includegraphics[width=13.63cm]{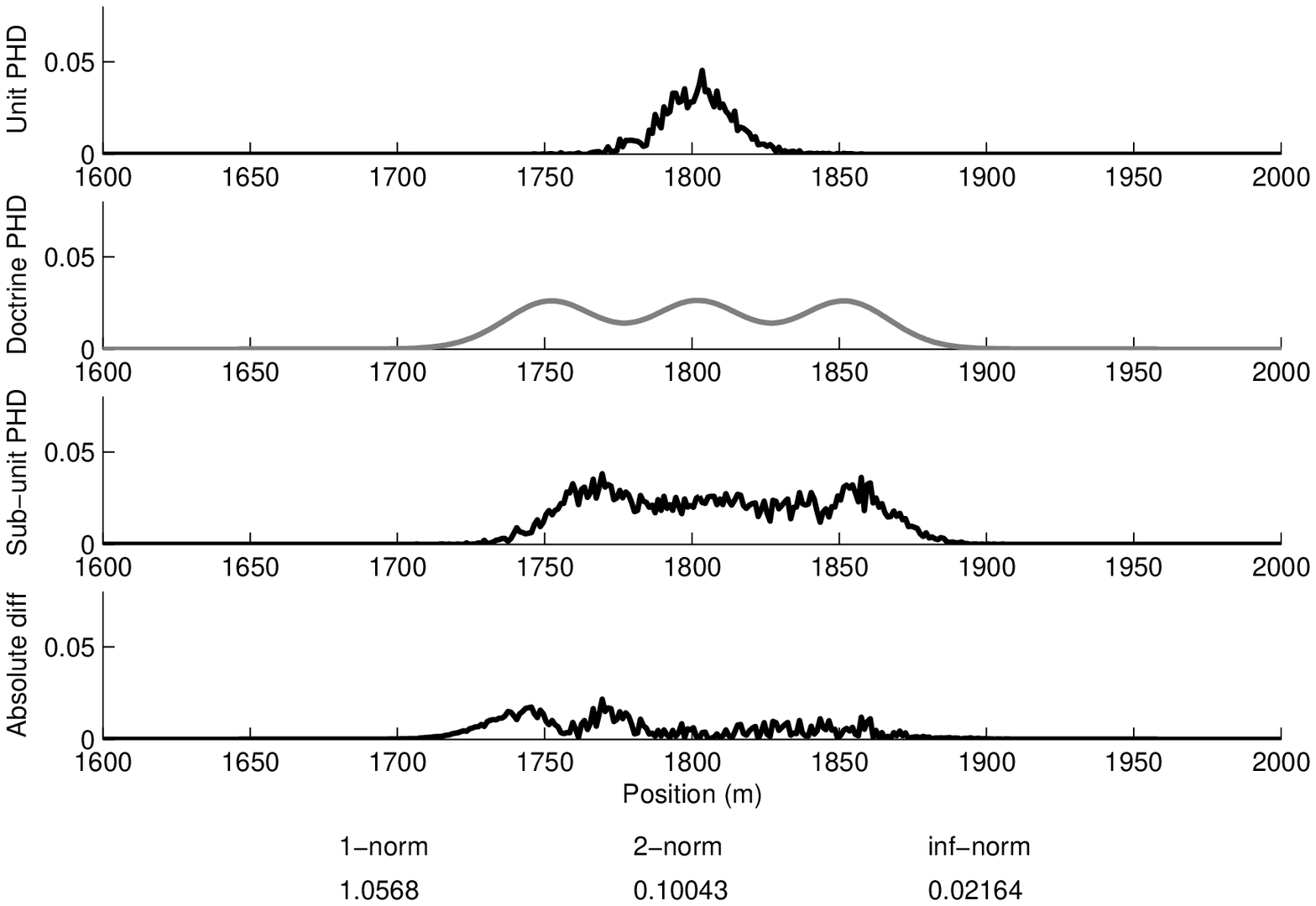}}}
\centerline{\subfigure[Very loose doctrine.]{\includegraphics[width=13.63cm]{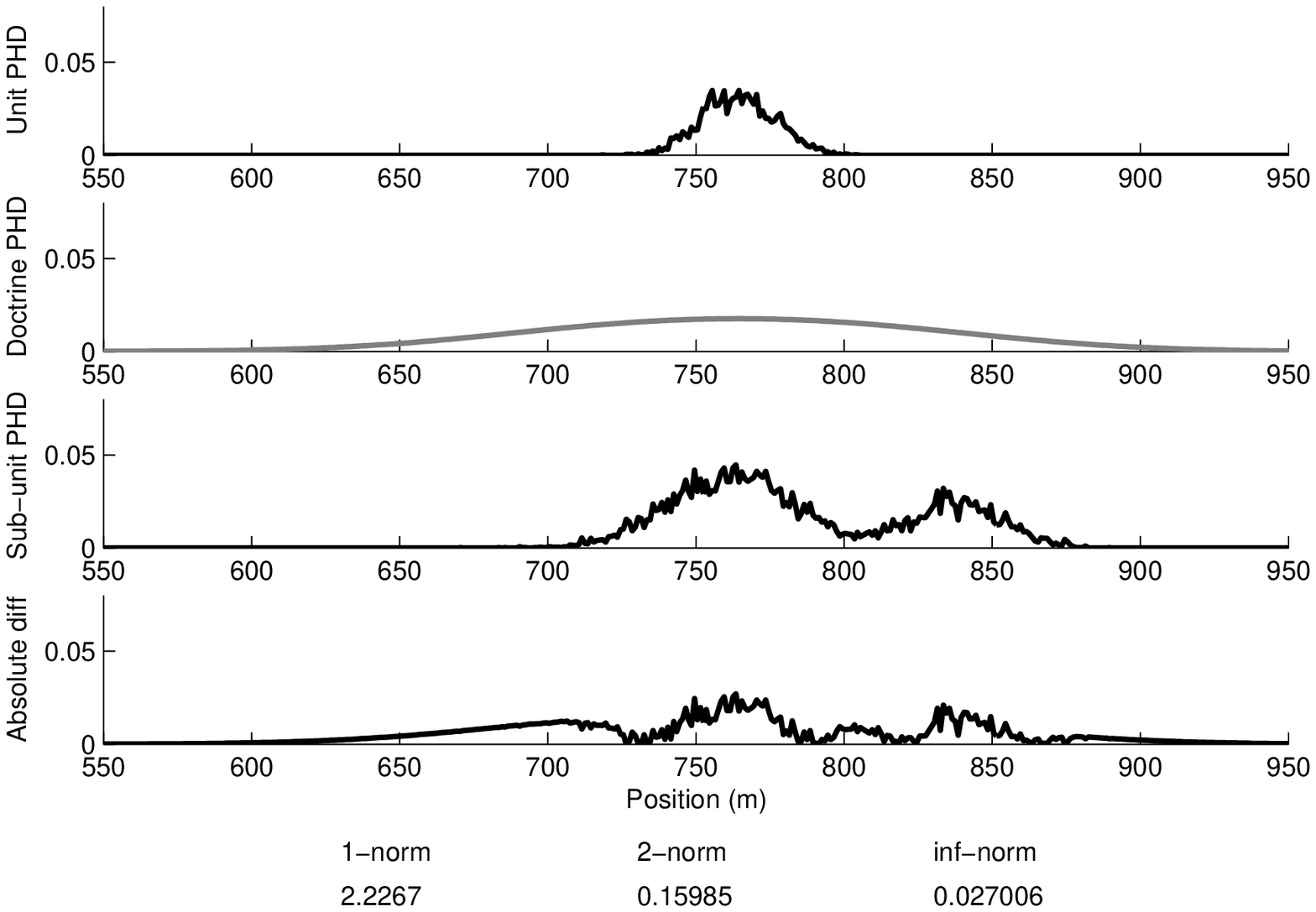}}}
\end{figure}

\subsection{Results}
The program was executed with three different values of 
$\sigma_{\mathit{Doctrine}}$. The graphs in Figures \ref{fig:experiments}a,b,c 
show, from top to bottom,  $D_{U,t}$, $D^*_{SU,t}$, $D_{SU,t}$ and 
$|D_{SU,t} - D^*_{SU,t}|$. The distances $d_1(D_{t,SU}, D^*_{t,SU})$, 
$d_2(D_{t,SU}, D^*_{t,SU})$ and $d_\infty(D_{t,SU}, D^*_{t,SU})$ are also shown
below the graphs.

Figure \ref{fig:experiments}a shows the result 
of a very exact doctrine, where $\sigma_{\mathit{Doctrine}} \rightarrow 0$. 
Since there is no randomness in the doctrine, the difference between 
$D^*_{SU,t}$ and $D_{SU,t}$ is only due to the observation noise and to 
approximations introduced by the particle filters. 

If $\sigma_{\mathit{Doctrine}}$ is on the same order of magnitude as the 
observation position noise $\sigma^x_O$ (Figure \ref{fig:experiments}b), the difference is on 
average the same as in case a. However, if 
$\sigma_{\mathit{Doctrine}} = x_{\mathit{Doctrine}}$ 
(Figure \ref{fig:experiments}c) the difference fluctuates greatly. Note that, 
since there are three sub-units in this example, a $d_1(D_{t,SU}, D^*_{t,SU}) 
\approx 6$ would indicate that the filters give totally different state 
estimates. In Figure \ref{fig:experiments}c, $d_1(D_{t,SU}, D^*_{t,SU}) = 2.3$. Thus, the norm measures still indicate a certain similarity between $D^*_{SU,t}$ and $D_{SU,t}$.

\section{Conclusions}
\label{sec:conclusions}

We presented a method for comparing the output of two different trackers, one 
tracking units of some type, the other tracking its sub-units. The output PHD 
from the unit tracker was transformed using doctrine knowledge, resulting in a 
``synthesized'' sub-unit PHD which could be compared to the actual sub-unit 
PHD.

The difference between the ``synthesized'' and the measured sub-unit PHD could 
then be used to alert a human in the loop or as an input to a sensor 
adaptation method. Possible uses of the difference is further discussed in the 
discussion below.

\subsection{Discussion}
\label{sec:discussion}

The most obvious area of application for our comparison method is failure 
detection. Imagine, e.g., a command and control system which has access to 
several fusion methods. Some are slow but reliable, others fast but more 
approximative. Potentially, they measure different things, in our example, 
the state of units and their sub-units. A signal of difference can then be 
used to alert the user of the system when the outputs of methods differ.

It should be note that a high difference can indicate any of several types of 
failure. One reason could be that one of the fusion methods are failing, in 
the case of tracking methods, that one of the methods lost track. However, the 
cause might also be that the doctrines used for transforming
PHD on different levels are wrong, or that the tracked targets (units and 
sub-units) suddenly stopped using them. It is important to draw the user's 
attention to both these types of events.

In principle, it is also possible to use
the information extracted from the difference
in order to do automatic
sensor adaptation to get more information
on what is happening in the affected area. This
should however be done with some care, since
the method must be extended significantly before it could
be used as the basis for any kind of automatic
reallocation of resources.

Sometimes it is not enough to get a global alert
that something unexpected is occurring
or that one of the fusion methods is malfunctioning.
An operator or a sensor allocation system might want to also know 
approximately where in the area of interest that the difference appears.
All the norm functions discussed in Sections \ref{sec:compare} and 
\ref{sec:experiments} were used globally. However, this is no principal
requirement that the norm computations are performed over the whole 
state-space. Thus, it is possible to apply them to small or big sub-areas.
An efficient failure localization algorithm could work as follows: After a 
global high difference alert is obtained, a simple search procedure is 
initiated. The state-space is divided into subparts. A local 
difference check is performed in each of these. High difference sub-areas are 
then searched recursively in the same way until a satisfactory resolution is 
obtained.

\subsection{Future work}
\label{sec:future}

We see several possibilities for future work related to the ideas
presented here.

The purpose of the method is to give the user of a decision support system
a warning when two different methods give
different results for the same problem.
Before this can be implemented, we need to determine to what extent
this is a desired feature from a user perspective, 
and if the feature can be presented to users
in a way that does not hinder their
process of obtaining a understanding of the situation.

Furthermore, the metrics used here are the simplest possible -- 
others could be investigated too.
It would also be interesting to compare
full joint multi-target probability density functions instead of PHD 
functions as in the current method.
In this case, the Wasserstein metric~\cite{hoffmanmahler02} could be used.

Another extension concerns the state-space. In this paper, we choose to use a 
1D scenario in order to make it easier to visualize the PHD's and their 
differences. The same ideas as are presented in this paper can be
applied to higher-dimensional scenarios. Naturally, a higher-dimensional 
state-space makes the problem of predicting doctrines more complex. Among 
other problems, it would be necessary to automatically determine 
which doctrine, out of several possible, is used. A possibility here would
be to add a doctrine-learning algorithm to the fusion module.

{\small
\bibliographystyle{spiebib}
\bibliography{ref}
}
\end{document}